\documentclass[a4paper,twoside]{article}

\usepackage{epsfig}
\usepackage{subfigure}
\usepackage{calc}
\usepackage{amssymb}
\usepackage{amstext}
\usepackage{amsmath}
\usepackage{multicol}
\usepackage{pslatex}
\usepackage{apalike}
\usepackage{fancyhdr}
\usepackage{scitepress}
\usepackage[small]{caption}

\usepackage{times}
\usepackage{graphicx}

\begin{document}
\onecolumn \maketitle \normalsize \vfill

\section{Introduction}
\label{sec:introduction}
Many different approaches have been proposed to track the motion of mobile objects in video \cite{Yilmaz}. However the tracking algorithm performance is always dependant on scene conditions such as illumination, occlusion frequence, movement complexity level. Some researches aim at improving the tracking quality by extracting the scene information such as: directions of paths, interesting zones. These elements can help the system to give a better prediction and decision on object trajectories. For example \cite{makris} have presented a method to model the paths in scenes based on detected trajectories. The system uses an unsupervised machine learning technique to compute trajectory clustering. A graph is automatically built to represent the path structure resulting from learning process. In \cite{dpchauGT}, the authors have proposed a global tracker to repair lost trajectories. The system learns automatically the ``lost zone'' where the tracked objects usually lose their trajectories and ``found zone'' where the tracked objects usually reappear. The system also takes complete trajectories to learn the common scene paths composed by $<$entrance zone, lost zone, found zone$>$. The learnt paths are then used to fuse the lost trajectories. This algorithm needs a 3D calibration environment and also a 3D person model as the inputs. These two papers get some good results but both require an off-line machine learning process to create rules for improving the tracking quality.
 
In order to solve the given problems in mobile object tracking, we propose in this paper a multiple feature tracker combining with a global tracking. We use first the Kalman filter to predict positions of tracked objects. However, this filter is only an estimator for linear movements while the object movements in surveillance videos are usually complex. A poor lighting condition of scene also influences to the tracking quality. Therefore, in this paper we propose to use different features to obtain more correct matching links between objects in a given time window. We also define a global tracker which does not require 3D environment calibration or off-line learning to improve tracking quality.

The rest of paper is organized as follows: The next section presents in detail the tracking process. Section 3 describes a global tracking algorithm which aims at filtering out noisy trajectories and fusing fragmented trajectories. This section also presents when a tracked object ends its trajectory. Section 4 shows in detail the results of the experimentation and validation. A conclusion is given in the last section as well as future work.

\section{Tracking Algorithm}
The proposed tracker takes as its input a bounding box list of detected objects at each frame. Pixel values inside these bounding boxes are also required to compute color metric. A tracked object at frame $t$ is represented by a state $s = [x, y, l, h] $ where ($x$, $y$) is center position, $l$ is width and $h$ is height of its 2D object bounding box at frame $t$. In the tracking process, we follow three steps of the Kalman filter: estimation, measurement and correction. However our contribution focus on the measurement step. The estimation step is first performed to estimate the new state of a tracked object in the current frame. The measurement step is then performed to search for the best detected object similar to each tracked object in the previous frames. The state of the found object refers to as ``measured state''. The correction step is finally performed to compute the ``corrected state'' of mobile object resulting from the ``estimated state'' and the ``measured state''. This state is considered as the official state of the considered tracked object in the current frame. For each detected object which does not match with any tracked object, a new tracked object with the same position and size will be created.

\subsection{Estimation of Position and Size}
For each tracked object in the previous frame, the Kalman filter is used to estimate the new state of the object in the current frame. The Kalman filter is composed of a set of recursive equations used to model and evaluate object linear movement. Let $s_{t-1}^+$ be the corrected state at instant $t-1$, the estimated state at time $t$, denoted $s_t^-$, is computed as follows:
\begin{equation}
s_t^- = \Phi s_{t-1}^+
\end{equation}

\noindent where $\Phi$ is the state transition matrix of $n$ x $n$ where $n$ is the considered feature number ($n = 4$ in our case). Note that in practice $\Phi$ might change with each time step, but here we assume it is constant. One of the drawbacks of the Kalman filter is the restrictive assumption of Gaussian posterior density functions at every time step, as
many tracking problems involve non-linear movement. In order to overcome this limitation, we give a weight value to determine the reliability of estimation computation and also of measurement (see section \ref{secCorrection} for details).

\subsection{Measurement}
This is our main contribution in the tracking process. For each tracked object in the previous frame, the goal of this step is to search for the best matched object in the current frame. In tracking problem, the execution time of tracking algorithm is very important to assure a real time system. Therefore, in this paper we propose to use a set of four features: distance, shape ratio, area and color histogram to compute the similarity between two objects. The computation of all of these features are not time consuming and the proposed tracker can thus be executed in real time. Because all measurements are computed in the 2D space, our proposed method does not require scene calibration information. For each feature $i$ ($i = 1..4$), we define a local similarity $LS_{i}$ in the interval $[0, 1]$ to quantify the object similarity of the feature $i$. A global similarity is defined as a combination of these local similarities. The detected object with the highest global similarity will be chosen for the correction step.

\subsubsection{Distance Similarity}
The distance between two objects is computed as the distance between the two corresponding object positions. Let $D_{max}$ be the possible maximal displacement of mobile object for $1$ frame in video and $d$ be the distance of two considered objects in two consecutive frames, we define a local similarity $LS_1$ between these two objects using distance feature as follows:
\begin{equation}
\label{eqDistance}
LS_1 = max (0,\ 1 - d/(D_{max} * m))
\end{equation}
\noindent where $m$ is the temporal difference (frame unity) of the two considered objects. 

In a 3D calibration environment, a value of $D_{max}$ can be set for the whole scene. However, this value should not be unique in a 2D scene. This threshold will change according to the distance between considered objects and the camera position. The nearer object to the camera, the larger its displacement is. In order to overcome this limitation, we set the $D_{max}$ value to the length half of bounding box diagonal of the considered tracked object.

\subsubsection{Area Similarity}
\label{secArea}
The area of an object $i$ is calculated by 
$W_iH_i$ where $W_i$ and $H_i$ are the 2D width and height
of the object respectively. A local similarity $LS_2$ between two areas of objects $i$ and $j$ is defined by:
\begin{equation}
LS_2 = \frac {min (W_iH_i,\ W_jH_j)} {max (W_iH_i,\ W_jH_j)} 
\end{equation}

\subsubsection{Shape Ratio Similarity}
The shape ratio of an object $i$ is calculated by 
$W_i/H_i$ (where $W_i$ and $H_i$ are defined in section \ref{secArea}). A local similarity $LS_3$ between two shape ratios of objects $i$ and $j$ is defined as follows:
\begin{equation}
LS_3 = \frac {min (W_i/H_i,\ W_j/H_j)} {max (W_i/H_i,\ W_j/H_j)} 
\end{equation}

\subsubsection{Color Histogram Similarity}
In this work, the color histogram of a mobile object is
defined as a histogram of pixel number inside its bounding box. Other color features (e.g. MSER) can be used but this one has given satisfying results. We define a local similarity $LS_4$ between two objects $i$ and $j$ for color feature as follows:
\begin{equation}
\label{eqColor}
LS_4 = \frac {\sum_{k = 1}^{n} rate_k } {n}
\enspace
\end{equation}

\noindent where $n$ is a parameter representing the number of histogram bins, $n = 1..768$ (the value 768 is the result of product 256 x 3) and $rate_k$ is computed as follows:
\begin{equation}
rate_k = \frac {min (H_i(k), H_j(k))} {max (H_i(k), H_j(k))}
\end{equation}

\noindent $H_i(k)$ and $H_j(k)$ are successively the number of pixels of object $i$, $j$ at bin $k$. There are some different ways to compute the difference between two histograms, in this work we choose the ratio computation for each histogram bin to obtain a value $rate_k$ normalised in the interval [0, 1]. Consequently the $LS_4$ value also varies in this interval.

\subsubsection{Global Similarity}
A detected object compared to previous frames can have some size variations because of detection errors or some color variations by illumination changes, but its maximum speed cannot exceed a determined value. Therefore in our work, the global similarity value takes into account a priority of distance feature compared to other features to decrease the number of false object matching links.
\begin{equation}
GS = \begin{cases}
	\frac{\sum_{i=1}^{4} w_iLS_i} {\sum_{j=1}^{4} w_j} & \mbox{if } LS_1 > 0 \\ \\
	0 	&	\mbox{otherwise}
      \end{cases}
\end{equation}

\noindent where $GS$ is the global similarity; $w_i$ is the weight (i.e. reliability) of feature $i$ and $LS_i$ is the local similarity of feature $i$. The detected object with the highest global similarity value $GS$ will be chosen as the matched object if:
\begin{equation}
\label{eqGS}
GS \geq T_1
\end{equation}

\noindent where $T_1$ is a predefined threshold. Higher the value of $T_1$ is set, more correct the matching links are established, but a too high value of $T_1$ can make lose the matching links in some complex environment (e.g. poor lighting condition, occlusion). The state of this object (including its position and its bounding box size) is called ``measured state''. At a time instant $t$, if a tracked object cannot find its matched object, the measured state $MS_t$ is set to $0$. In the experimentation of this work, we suppose that all feature weight $w_i$ have the same values.

\subsection{Correction}\label{secCorrection}
Thanks to the estimated and measured states, we can update the position and size of tracked object by computing the corrected state as follows:
\begin{equation}
CS_t = \begin{cases}
	wMS_t + (1 - w)ES_t & \mbox{if } MS_t \neq 0 \\ \\
	MS_{t-1}	& \mbox{otherwise}
	\end{cases}
\end{equation}  

\noindent where $CS_t$, $MS_t$, $ES_t$ are the corrected state, measured state and estimated state of the tracked object at time instant $t$ respectively; $w$ is the weight of measurement state. If the measured state is not found, the corrected state will be set equal to the corrected state in the previous frame. While the estimated state is only result of a simple linear estimator, the measurement step is fulfilled by considering four different features. We thus set a high value to $w$ ($w = 0.7$) in our experimentation.

\section{Global Tracking Algorithm}
Global tracking aims at fusing the fragmented trajectories belonging to a same mobile object and removing the noisy trajectories. As mentioned in section \ref{secCorrection}, if a tracked object cannot find the corresponding detected object, his corrected state will be set to the current corrected state. The object then turns into a ``waiting state''. This tracked object goes out of ``waiting state'' when it finds its matched object. A tracked object can turn into and go out of ``waiting state'' many times during its life. This waiting step allows us to let a non-updated tracks live for some frames when no correspondence is found. The system can so track completely object motion even when the object is not sometime detected or is detected incorrectly. This prevents the mobile object trajectories from being fragmented. However, the ``waiting state'' can cause an error when the corresponding mobile object goes out of the scene definitively. Therefore, we propose a rule to decide the moment when a tracked object ends its life and also to avoid maintaining for too long the ``waiting state''. A more reliable tracked object will be kept longer in the ``waiting state''. In our work, the tracked object reliability is directly proportional to number of times this object finds matched objects. The greater number of matched objects, the greater tracked object reliability is. Let Id of a frame be the order of this frame in the processed video sequence, a tracked object ends if:
\begin{equation}
\label{finished}
F_l < F_c - min(N_r, T_2)
\end{equation}

\noindent where $F_l$ is the latest frame Id where this tracked object finds matched object (i.e. the frame Id before entering the ``waiting state''), $F_c$ is the current frame Id, $N_r$ is the number of frames in which this tracked object was matched with a detected object, $T_2$ is a parameter to determine the number of frames for which the ``waiting state'' of a tracked object cannot exceed. With this calculation method, a tracked object that finds a greater number of matched objects is kept in the ``waiting state'' for a longer time but its ``waiting state'' time never exceed $T_2$. Higher the value of $T_2$ is set, higher the probability of finding lost objects is, but this can decrease the correctness of the fusion process.

We also propose a set of rules to detect the noisy trajectories. The noise usually appears when wrong detection or misclassification (e.g. due to low image quality) occurs. A static object or some image regions can be detected as a mobile object. However, a noise usually only appears in few frames or does not displace really (around a fixed position). We thus propose to use temporal and spatial filters to remove it. A trajectory is composed of objects throughout time, so it is unreliable if it cannot contain enough objects and usually lives in the ``waiting state''. Therefore we define a temporal threshold when a ``waiting state'' time is greater, the corresponding trajectory is considered as noise. Also, if a new trajectory appears, the system cannot determine immediately whether it is noise or not. The global tracker has enough information to filter out it only after some frames since its appearance moment. Consequently, a trajectory that satisfies one of the following conditions, is considered as noise:
\begin{equation} 
\label{condTL}
	T < T_3
\end{equation}
\begin{equation} 
\label{condDmax}
	(d_{max} < T_4) \mbox{ } and \mbox{ } (T \geq T_3)
\end{equation}
\begin{equation} 
\label{condWTL}
	(\frac{T_w}{T} \geq T_5) \mbox{ } and \mbox{ } (T \geq T_3)
\end{equation}

\noindent where $T$ is time length (number of frames) of the considered trajectory (``waiting state'' time included); $d_{max}$ is the maximum spatial length of this trajectory; $T_w$ is the total time of ``waiting state'' during the life of the considered trajectory; $T_3$, $T_4$ and $T_5$ are the predefined thresholds. While $T_4$ is a spatial filter threshold, $T_3$ and $T_5$ can be considered as temporal filter thresholds to remove noisy trajectories. The condition (\ref{condTL}) is only examined for the trajectories which end their life according to equation (\ref{finished}).

\section{Experimentation and Validation}
\begin{figure*}[t]
\centering
\includegraphics[width=16cm]{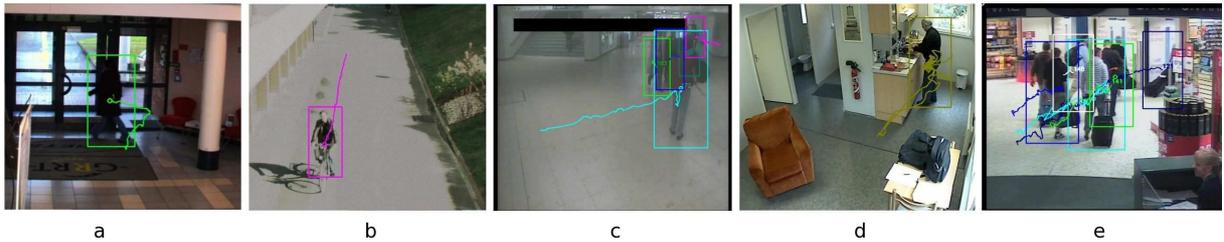}
\caption{Illustration of tested videos: a) ETI-VS1-BE-18-C4 b) ETI-VS1-RD-16-C4 c) ETI-VS1-MO-7-C1 d) Gerhome e)\~ TRECVid. The colors represent the bounding boxes and trajectories of tracked people.}
\label{figCom}
\end{figure*}
We can classify the tracker evaluation methods by two principal approaches: off-line evaluation using ground truth data \cite{needham} and on-line evaluation without ground truth data \cite{dpchauEva}. In order to be able to compare our tracker performance with the other ones, we decide to use the tracking evaluation metrics defined in ETISEO benchmarking project \cite{atnghiem} which comes from the first approach. The first tracking evaluation metric $M_1$, which is the ``tracking time'' metric measures the percentage of time during which a reference object (ground truth data) is tracked. The second metric $M_2$ ``object ID persistence'' computes throughout time how many tracked objects are associated with one reference object. The third metric $M_3$ ``object ID confusion'' computes the number of reference object IDs per tracked object. These metrics must be used together to obtain a complete tracker evaluation. Therefore, we also define a tracking metric $\overline{M}$ taking the average value of these three tracking metrics. All of the four metric values are defined in the interval [0, 1]. The higher the metric value is, the better the tracking algorithm performance gets.

In this experimentation, we use the people detection algorithm based on HOG descriptor of the OpenCV library (http://opencv.willowgarage.com/wiki/). Therefore we focus the experimentation on the sequences containing people movements. However the principle of the proposed tracking algorithm is not dependent on tracked object type.

We have tested our tracker with five video sequences. The first three videos are selected from ETISEO data in order to compare the proposed tracker performance with that from other teams. The last two videos are extracted from different projects so that the proposed tracker can be tested with more scene conditions. All of these five videos are tested with the following parameter values: $n = 96$ bins (formula (\ref{eqColor})), $T_1 = 0.8$ (formula (\ref{eqGS})), $T_2 = 20$ frames (formula (\ref{finished})), $T_3 = 20$ frames	(formula (\ref{condTL})), $T_4 = 5$ pixels (formula (\ref{condDmax})) and $T_5 = 40\%$ (formula (\ref{condWTL})).

The first tested ETISEO video shows a building entrance, denoted ETI-VS1-BE-18-C4. In this sequence, there is only one person moving, but the illumination and contrast level are low (see image a of figure \ref{figCom}). The second ETISEO video shows a road with strong illumination, denoted ETI-VS1-RD-16-C4 (see image b of figure \ref{figCom}). There are walker, bicyclists, car moving on the road. The third video shows an underground station denoted ETI-VS1-MO-7-C1 where there are many complex people movements (see image c of figure \ref{figCom}). The illumination and contrast in this sequence are very bad.

In this experimentation, tracker results from seven different teams in ETISEO have been presented: 1, 8, 11, 12, 17, 22, 23. Table \ref{tabRes} presents performance results of our tracker and of the ones of seven teams on three ETISEO sequences. Although each tested video has its proper complex, the tracking evaluation metrics of the proposed tracker get the highest values in most cases compared to other teams. In the second video, the tracking time of our tracker is low ($M_1 = 0.36$) because as mentioned above, we only use the people detector and so system usually fails to detect cars.
\begin{table}[t]
\fontsize{9}{10}\selectfont
	\begin{center}
		\begin{tabular}{|p{1.15 cm}|p{0.5 cm}|p{1.3 cm}|p{1.3 cm}|p{1.3 cm}|}

 		\hline
			& 	& \begin{scriptsize}ETI-VS1-BE-18-C4\end{scriptsize} & \begin{scriptsize}ETI-VS1-RD-16-C4\end{scriptsize}	& \begin{footnotesize}ETI-VS1-MO-7-C1\end{footnotesize} \\
		\hline
		&$N$		&1108	 & 1315 & 2282 \\
		\hline
			&$F$	&25 $fps$& 16 $fps$& 25 $fps$\\

		\hline
			 	&$M_1$	&\textbf{0.64}	&$0.36$	&\textbf{0.87}	\\
		\cline{2-5}
		  Proposed 	&$M_2$	&\textbf{1}	&\textbf{1} 	&\textbf{0.92}\\
		\cline{2-5}
		  tracker	&$M_3$ 	&\textbf{1}	&\textbf{1} 	&\textbf{1}\\
		\cline{2-5}
			& $\overline{M}$ &\textbf{0.88}	&\textbf{0.79} 	&\textbf{0.93}\\
 		\cline{2-5}
		& $s$  		&292 $fps$ &641 $fps$	&84 $fps$ \\

		\hline \hline
			 	&$M_1$	&$0.48$	&$0.44$	&$0.77$	\\
		\cline{2-5}
		  Team	 	&$M_2$	&$0.80$	&$0.81$	&$0.78$\\
		\cline{2-5}
		  1		&$M_3$ 	&$0.83$	&$0.61$	&\textbf{1}\\
		\cline{2-5}
		&$\overline{M}$	&$0.70$	&$0.62$	&$0.85$\\
		\hline	\hline		 	
				&$M_1$	&$0.49$	&$0.32$	&$0.58$	\\
		\cline{2-5}
		  Team	 	&$M_2$	&$0.80$	&$0.62$ &$0.39$\\
		\cline{2-5}
		  8		&$M_3$ 	&$0.77$	&$0.52$	&\textbf{1}\\
		\cline{2-5}
		&$\overline{M}$ &$0.69$	&$0.49$	&$0.66$\\
		\hline	\hline		 	
				&$M_1$	&$0.56$	&\textbf{0.53} &$0.75$	\\
		\cline{2-5}
		  Team	 	&$M_2$	&$0.71$	&$0.94$	&$0.61$\\
		\cline{2-5}
		  11		&$M_3$ 	&$0.77$	&$0.81$	&$0.75$\\
		\cline{2-5}
		&$\overline{M}$ &$0.68$	&$0.76$ &$0.70$\\
		\hline	\hline		 	
				&$M_1$	&$0.19$	&$0.40$	&$0.58$	\\
		\cline{2-5}
		  Team	 	&$M_2$	&\textbf{1}	&\textbf{1} 	&$0.39$\\
		\cline{2-5}
		  12		&$M_3$ 	&$0.33$	&$0.83$ &\textbf{1}\\
		\cline{2-5}
		& $\overline{M}$ &$0.51$	&$0.74$	&$0.91$\\
		\hline	\hline		 	
				&$M_1$	&$0.17$	&$0.35$	&$0.80$	\\
		\cline{2-5}
		  Team	 	&$M_2$	&$0.61$	&$0.81$	&$0.57$\\
		\cline{2-5}
		  17		&$M_3$ 	&$0.80$	&$0.66$	&$0.57$\\
		\cline{2-5}
		& $\overline{M}$ &$0.53$	&$0.61$	&$0.65$\\
		\hline	\hline		 	
				&$M_1$	&$0.26$	&$0.36$	&$0.78$	\\
		\cline{2-5}
		  Team	 	&$M_2$	&$0.35$	&$0.43$	&$0.36$\\
		\cline{2-5}
		  22		&$M_3$ 	&$0.33$	&$0.20$	&$0.54$\\
		\cline{2-5}
		& $\overline{M}$ &$0.31$	&$0.33$	&$0.56$\\
		\hline	\hline		 	
				&$M_1$	&$0.05$	&$0.03$	&$0.05$	\\
		\cline{2-5}
		  Team	 	&$M_2$	&$0.46$	&$0.73$	&$0.61$\\
		\cline{2-5}
		  23		&$M_3$ 	&$0.39$	&$0.23$	&$0.42$\\
		\cline{2-5}
		& $\overline{M}$ &$0.30$	&$0.33$	&$0.36$\\
		\hline
		\end{tabular}
		\caption{\label{tabRes}Summary of tracking results for ETISEO videos. $N$: video frame number, $F$: video frame rate, $s$ : average processing speed of the tracking task ($frames/second$) (not taking into account the detection process). The highest values are printed bold.}
	\end{center}
\end{table}

The fourth video sequence has been provided by the Gerhome project (see image d of figure \ref{figCom}). The objective of this project is to enhance autonomy of the elderly people at home by using intelligent technologies for house automation. In this sequence, there is only one person moving but the video length is quite long (13 minutes 40 seconds). We can find tracking results in the second column of table \ref{tabRes0}. Although the sequence length is quite long, the proposed tracker can follow person movement for most of the time, from frame 1 to frame 8807 ($M_1 = 0.86$). After that, there are four moments when the detection algorithm cannot detect the person in an interval over 20 frames (over the value of $T_2$). Therefore the value of metric $M_2$ for this video sequence is only equal to $0.2$.

The last tested sequence concerns the movements of people in an airport. This sequence is provided by TREC Video Retrieval Evaluation (TRECVid) \cite{trecvid}. The people tracking in this sequence is a very hard task because there are always a great number of movements in the scene and occlusions usually happen (see image e of figure \ref{figCom}). Despite these difficulties, the proposed tracker obtains high values for all three tracking evaluation metrics: $M_1 = 0.71$, $M_2 = 0.90$ and $M_3 = 0.85$ (see the third column of table \ref{tabRes0}).
\begin{table}[h]
\fontsize{9}{10}\selectfont
	\begin{center}
		\begin{tabular}{|p{3.0cm}|p{1.5cm}|p{1.5cm}|}
 		\hline
				 	&Gerhome & TRECVid \\	
		\hline
			Number of frames&10240	 & 5000 \\
		\hline
			Frame rate	&12 $fps$ & 25 $fps$ \\
		\hline

			 	$M_1$	&0.86	&0.71	\\
		\hline
		  		$M_2$	&0.20	&0.90 	\\
		\hline
		  		$M_3$ 	&1	&0.85 	\\
		\hline
			$\overline{M}$ 	&0.69	&0.82 	\\
		\hline
				$s$	&$58$ $fps$ &$20$ $fps$ \\
		\hline
		\end{tabular}
		\caption{\label{tabRes0}Tracking results for Gerhome and TRECVid videos. $s$ denotes the average processing speed of the tracking task ($frames/second$).}
	\end{center}
\end{table}

The average processing speed of the proposed tracking algorithm for all considered sequences is very high. In the most complicated sequence where there are many crowds (TRECVid sequence), this value is equal to $20\ fps$. In the other video sequences, the average processing speed of the tracking task is greater than $50\ fps$. This helps whole tracking framework (including video acquisition, detection and tracking tasks) can become a real time system.

\section{Conclusion}
Although many researches aim at resolving the problems given by tracking process such as misdetection, occlusion, there is still not a robust tracker which can well perform in different scene conditions. This paper has presented a tracking algorithm which is combined with a global tracker to increase the robustness of the tracking process. The proposed approach has been tested and validated on five real video sequences. The experimentation results show that the proposed tracker can obtain good tracking results in many different scenes although each tested scene has its proper complexity. Our tracker also gets the best performances in the experimented ETISEO videos compared to other tracker evaluated in this project. The average processing speed of the proposed tracking algorithm is high. However, some drawbacks still exist in this approach: the used features are simple, more complex features (e.g. color covariance) are needed to obtain the more reliable matching links between objects. We also propose in future work an on-line automatic learning of the detected trajectories to improve the global tracker quality.

\section*{\uppercase{Acknowledgements}}
\noindent This work is supported by The PACA region, The General Council of Alpes Maritimes province, France as well as The ViCoMo, Vanaheim, Video-Id, Cofriend and Support projects.

%

\bibliographystyle{apalike}
{\small
\bibliography{example}}
\renewcommand{\baselinestretch}{1}

%

\end{document}